\documentclass[conference]{IEEEtran}
\IEEEoverridecommandlockouts

\usepackage{cite}
\usepackage{amsmath,amssymb,amsfonts}
\usepackage{algorithmic}
\usepackage{graphicx}
\usepackage{textcomp}
\usepackage{xcolor}
\usepackage{caption}
\usepackage{subcaption}
\def\BibTeX{{\rm B\kern-.05em{\sc i\kern-.025em b}\kern-.08em
    T\kern-.1667em\lower.7ex\hbox{E}\kern-.125emX}}

\let\OLDthebibliography\thebibliography
\renewcommand\thebibliography[1]{
  \OLDthebibliography{#1}
  \setlength{\parskip}{0pt}
  \setlength{\itemsep}{0pt plus 0.3ex}
}

\begin{document}

\def\x{{\mathbf x}}
\def\L{{\cal L}}

\title{Q-YOLOP: Quantization-aware You Only Look Once for Panoptic Driving Perception}


\author{
\IEEEauthorblockN{
Chi-Chih Chang\IEEEauthorrefmark{1}\textsuperscript{1}, 
Wei-Cheng, Lin\IEEEauthorrefmark{1}\textsuperscript{1},
Pei-Shuo Wang\IEEEauthorrefmark{1}\textsuperscript{1},
Sheng-Feng Yu\IEEEauthorrefmark{1}\textsuperscript{1}\textsuperscript{2}, \\
Yu-Chen Lu\IEEEauthorrefmark{1}\textsuperscript{1}\textsuperscript{2},
Kuan-Cheng Lin\IEEEauthorrefmark{1}\textsuperscript{1} and
Kai-Chiang Wu\textsuperscript{1}}
\IEEEauthorblockA{
    \textsuperscript{1} National Yang Ming Chiao Tung University\\
    \textsuperscript{2} Macronix International Co., Ltd.}
}

\maketitle
\begin{abstract}
In this work, we present an efficient and quantization-aware panoptic driving perception model (Q-YOLOP) for object detection, drivable area segmentation, and lane line segmentation, in the context of autonomous driving. Our model employs the Efficient Layer Aggregation Network (ELAN) as its backbone and task-specific heads for each task. We employ a four-stage training process that includes pretraining on the BDD100K dataset, finetuning on both the BDD100K and iVS datasets, and quantization-aware training (QAT) on BDD100K. During the training process, we use powerful data augmentation techniques, such as random perspective and mosaic, and train the model on a combination of the BDD100K and iVS datasets. Both strategies enhance the model’s generalization capabilities. The proposed model achieves state-of-the-art performance with an mAP@0.5 of 0.622 for object detection and an mIoU of 0.612 for segmentation, while maintaining low computational and memory requirements.
\end{abstract}

\begin{IEEEkeywords}
Object detection, semantic segmentation, quantization-aware training, autonomous driving
\end{IEEEkeywords}
\section{Introduction}
\label{sec:intro}

Panoptic perception systems are critical components of autonomous cars, enabling them to perceive and understand their environment comprehensively. These systems solve multiple vision tasks simultaneously, including object detection, lane line segmentation, drivable area segmentation, and generate a rich understanding of the road scene. 

In order to solve the multi-task problem for panoptic driving perception, we develop a low-power, multi-task model tailored for traffic scenarios, addressing the challenges of object detection and semantic segmentation. The aim is to create efficient algorithms capable of accurately recognizing objects and segmenting both lane line and drivable area while maintaining minimal computational cost, rendering them ideal for deployment in resource-constrained environments such as mobile devices, IoT devices, and embedded systems.

\begin{figure}[t!]
    \includegraphics[width=\linewidth]{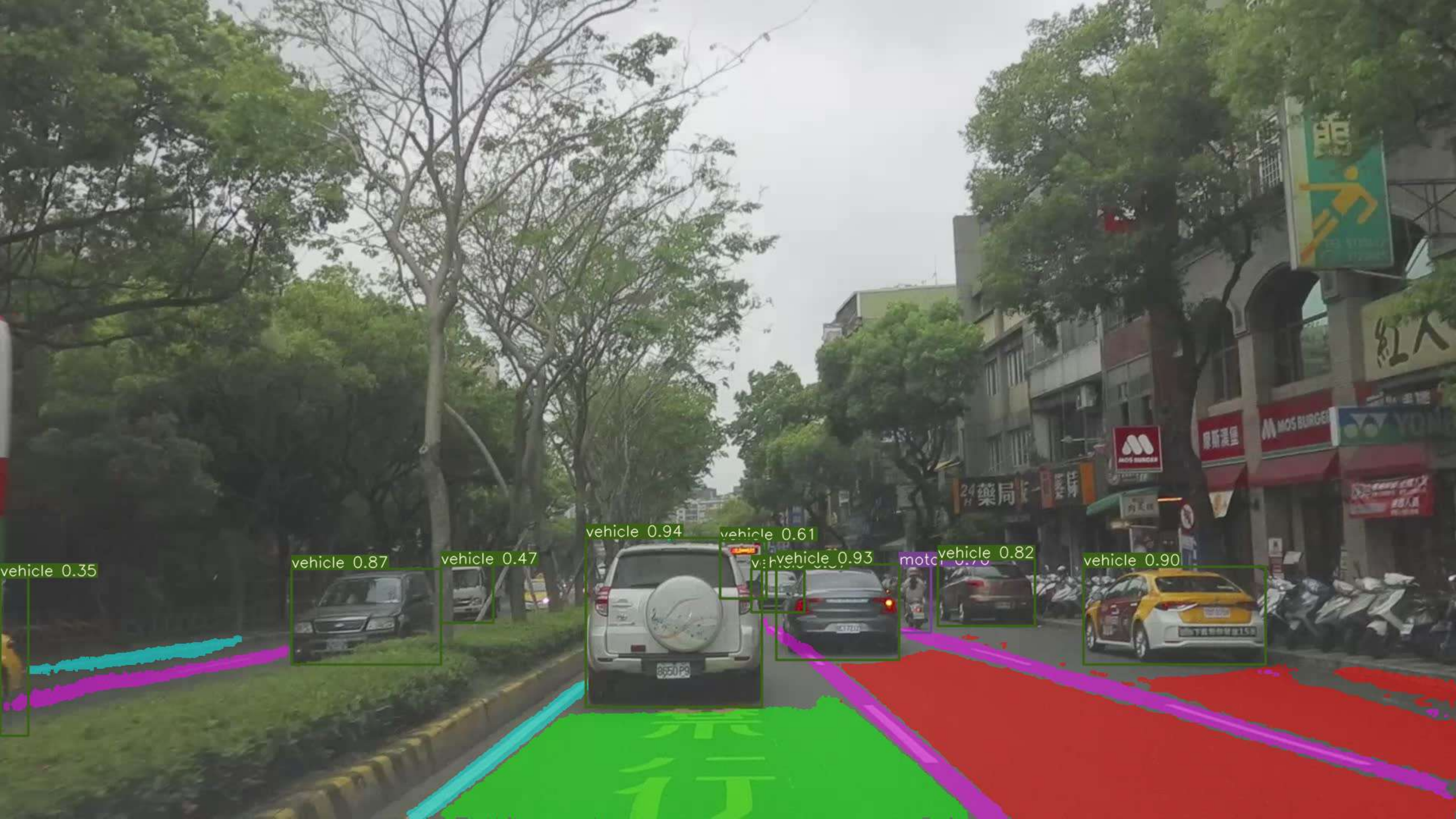}
    \caption{Our model is designed to simultaneously process object detection, drivable area segmentation, and lane line segmentation on a single input image. The bounding boxes indicate the location of traffic objects, the green areas represent the main lane of drivable areas, the red areas represent the alternate lane of drivable areas, the light blue areas represent single lines, and the pink-purple areas represent dashed lines.} 
\end{figure}


To achieve low-power consumption, we adopt a neural network architectures optimized for energy efficiency. The development process involves reducing the size and complexity of the models used for object detection and segmentation, as well as quantizing the model to minimize energy consumption.


Our panoptic driving perception system reaches $93.46$ FPS on NVIDIA V100 and $3.68$ FPS on MediaTek Dimensity 9200 Series Platform. Meanwhile, it attains $0.622$ mAP and $0.612$ mIoU on the object detection and segmentation tasks of the competition iVS dataset.

\section{Method}

Our model, derived from YOLOPv2~\cite{yolopv2} and YOLOv7~\cite{yolov7}, is specifically designed to address both object detection and segmentation tasks. It comprises five main components: the backbone, the neck, the detection head, drivable area segmentation head, and lane line segmentation head. The backbone is Efficient Layer Aggregation Network (ELAN)~\cite{ELAN}, optimized for rapid and efficient feature extraction.

The neck of our model is a Spatial Pyramid Pooling (SPP) network~\cite{spp}, which facilitates the handling of objects with varying scales and sizes by pooling features at multiple resolutions. This enhancement improves the accuracy and robustness of object detection. The detection head is based on RepConv~\cite{repvgg}, an innovative neural network architecture that merges the efficiency of mobile networks with the accuracy of more complex models. 
Subsequently, a non-maximum suppression is applied to the output of object detection process to generate the final predictions.
Consequently, our model is capable of accurately detecting objects in images while managing computation and memory requirements.

Furthermore, in addition to object detection, our neural network also encompasses task-specific heads for drivable area segmentation and lane line segmentation. These dedicated heads possess distinct network structures that are optimized for their respective tasks. As drivable area segmentation and lane line segmentation generate separate predictions, we allow the result of lane line segmentation to overlap with the result of drivable area segmentation.

\begin{figure}
    \centering
    \includegraphics[width=\linewidth]{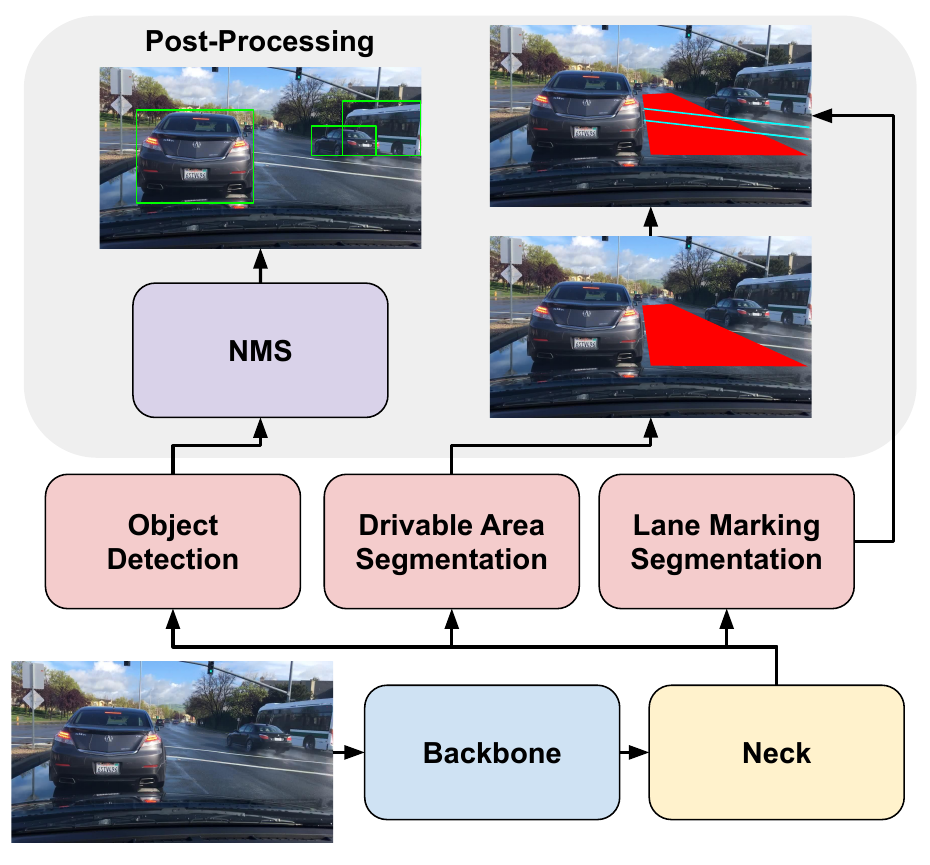}
    \caption{The proposed model architecture and post-processing flow. First, a non-maximum suppression (NMS) technique is applied to the output of the object detection head in order to refine the predictions. Moreover, the prediction of lane line segmentation is allowed to overwrite the prediction of drivable area segmentation in regions where both predictions overlap.}
    \label{fig:architecture}
\end{figure}

In summary, our model is engineered to optimize efficiency and accuracy while also addressing the challenges associated with multi-task. Its unique combination of components and specialized task heads make it ideal for real-world applications such as autonomous driving and object recognition in resource-constrained environments. A visual representation of our model architecture is presented in Figure~\ref{fig:architecture}.

\subsection{Loss Function}

As we modify the head of YOLOPv2~\cite{yolopv2} to support multi-label prediction, we introduce the loss function derived from HybridNets~\cite{vu2022hybridnets} to enhance the performance of our approach. The loss function for objection detection task consists of three components, 
\begin{equation}
  L_{det} = \alpha_1 L_{class} + \alpha_2 L_{obj} + \alpha_3 L_{box}  
\end{equation}

Specifically, for $L_{det}$, focal loss is used in both $L_{class}$ and $L_{obj}$. The classification loss, $L_{class}$, is responsible for penalizing classification errors, while  $L_{obj}$ is used for predicting object confidence. Both terms are implemented by focal loss~\cite{lin2017focal}. The term $L_{box}$ represents the similarity between the predicted results and ground truth by considering the overlap rate, aspect ratio, and scale. We implement $L_{box}$ using the smooth L1 loss function. The coefficient $\alpha_1$, $\alpha_2$, and $\alpha_3$ are hyperparameters used to balance the detection losses.

The objective for lane line segmentation task combines three components,
\begin{equation}
  L_{seg\_ll} = \beta_1 L_{Tversky} + \beta_2 L_{Focal} + \beta_3 L_{Jaccard}
\end{equation}

The first term Tversky loss~\cite{tversky}, $L_{Tversky}$, is used to address the issue of data imbalance and achieve much better trade-off between precision and recall, and the second term $L_{Focal}$ aims to minimize the classification error between pixels and focuses on hard labels. The third term, $L_{Jaccard}$, is utilized to measure the similarity between prediction and ground-truth segmentation masks.
The coefficient $\beta_1$, $\beta_2$ and $\beta_3$ are hyperparameters used to balance losses.

On the other hand, the objective for drivable area segmentation task only combines two components:
\begin{equation}
  L_{seg\_da} = \gamma_1 L_{Tversky} + \gamma_2 L_{Focal}
\end{equation}

The coefficient $\gamma_1$ and $\gamma_2$ are hyperparameters used to balance the losses.

The overall objective, $L_{all}$, for our final model combines the object detection loss $L_{det}$ and the segmentation loss $L_{seg}$ to learn both tasks at the same time:

\begin{equation}
  L_{all} = \delta_1 L_{det} + \delta_2 L_{seg\_da} + \delta_3 L_{seg\_ll}
\end{equation}

The coefficient $\delta_1$, $\delta_2$ and $\delta_3$ are hyperparameters used to balance the detection loss and segmentation losses.

\subsection{Quantization}
Quantization-Aware Training (QAT) is a technique aimed at making neural networks more amenable to quantization. During QAT, we introduce the quantization error during training by sequentially applying quantize and dequantize operations. This enables the network to learn more robust representations that can be efficiently quantized during inference.
We employ the Straight-Through Estimator (STE)~\cite{Jacob_2018_CVPR} algorithm for QAT, which offers a simple and efficient approach. With STE, we round the weights and activations to the nearest quantization level during forward propagation, while utilizing the gradients of the unquantized values during backward propagation. In this manner, the network can backpropagate the gradients through the quantization operation, which is not differentiable in its original form. By simulating the quantization error during training, we can ensure that the network learns robust features that are less sensitive to quantization.

\section{Implementation Detail}

\subsection{Data Preparation}
As the organizers of the contest provided only a portion of the BDD100K~\cite{Yu2018BDD100KAD} dataset, we opted to use the complete BDD100K dataset to augment the training data. In previous works that used the BDD100K dataset for semantic segmentation, the focus was typically on segmenting only the drivable areas and lane lines. There were no attempts to further classify the drivable areas or lane lines into multiple categories.

However, our semantic segmentation task involves categorizing images into six classes: background, main lane, alternative lane, single line, double line, and dashed line. This is different from previous works, which only segmented images into two classes: line and lane. Therefore, we re-generate the six classes of segmentation labels for the BDD100K dataset.

For the object detection task, the objective is to detect four types of objects: pedestrian, vehicle, scooter, and bicycle. In the case of scooters and bicycles, both the rider and the respective vehicle are included within the bounding box. However, the BDD100K dataset labels riders, scooters, and bicycles as distinct entities, as depicted in the following figure. To comply with the task requirements, we employ the Hungarian algorithm~\cite{Kuhn1955Hungarian} to pair riders with their corresponding scooters or bicycles and label them within the same bounding box.

\subsection{Training Process}
In our experiments, the training process consists of several stages: 1) initial pretraining on the BDD100K~\cite{Yu2018BDD100KAD} dataset, then 2) pretraining on the BDD100K with mosaic augmentation~\cite{Bochkovskiy2020YOLOv4OS}, 3) finetuning on both BDD100K and iVS datasets, 4) quantization-aware training (QAT) on the integrated iVS and BDD100K datasets. Initially, we train our model on the BDD100K dataset without mosaic for 300 epochs, then turning on mosaic augmentation for 150 epochs. Subsequently, we jointly train the model on both the BDD100K and iVS datasets for an additional 150 epochs. Finally, we apply QAT~\cite{Jacob_2018_CVPR} for an extra 20 epochs for quantization.


\textbf{Data Augmentation Techniques.} To enhance the model's generalization capabilities, we apply several data augmentation techniques during the training process. These techniques include normalization, random perspective transformation, HSV color space augmentation, horizontal flipping, and mosaic. By simulating variations that may occur in real-world scenarios, these techniques improve the model's ability to adapt to new data. The mosaic technique turns on in the second and third stages, and it is turned off for the last 10 epochs of third stage.

In detail, all images is normalized with mean $(0.485, 0.456, 0.406)$ and std $(0.229, 0.224, 0.225)$, random perspective transforming with scale factor $0.25$, and translation factor $0.1$. For HSV color space augmentation, the factor of Hue augmentation is $0.015$, the factor of Saturation augmentation is $0.7$, and the factor of Value augmentation is $0.4$.

\textbf{Weight Initialization.} The weight of the backbone and detection head of our model is initialized from YOLOv7~\cite{yolov7} pretrained weight, while the other parameters are all random initialized.

\textbf{Implementation Details.} We resize all images to $384 \times 640$ of both BDD100K~\cite{Yu2018BDD100KAD} and iVS datasets. The Adam optimizer is used for optimization. Different batch sizes are used for different stages, with $32$ during first and second pretraining, $32$ during finetuning, and $16$ during quantization-aware training (QAT). The default anchor sizes are set as (12,16), (19,36), (40,28), (36,75), (76,55), (72,146), (142,110), (192,243), and (459,401). The learning rate scheduler employed is cosine annealing with a warm-up phase, and the initial learning rates are set to 1e-2 during first pretraining, 5e-3 during second pretraining, 5e-4 during finetuning, and 5e-5 during QAT. The minimum learning rates are set to 1e-5 during first pretraining, 5e-6 during second pretraining, 5e-7 during finetuning, and 5e-8 during QAT. The warm-up phase is set to 5 epochs during pretraining and 0 epochs during finetuning and QAT. The values of the coefficients for the losses are reported as follows: $\alpha_1$ = 0.5, $\alpha_2$ = 1.0, $\alpha_3$ = 0.05, $\beta_1$ = 1.0, $\beta_2$ = 1.0, $\beta_3$ = 1.0, $\delta_1$ = 1.0, $\delta_2$ = 1.0, $\gamma_1$ = 0.2, $\gamma_2$ = 0.2, and $\gamma_3$ = 0.2. These coefficients are used in the computation of the loss function, which is a crucial component of our proposed method.

\subsection{Inference Process}
The inference process involves pre-processing the input images, which includes resizing from $1080 \times 1920$ to $384 \times 640$. Following this, images are normalized with mean $(0.485, 0.456, 0.406)$ and standard deviation $(0.229, 0.224, 0.225)$. The post-processing steps for the detection and segmentation parts are carried out. In the detection part, the intersection over union (IoU) threshold of non-maximum suppression (NMS) is set to $0.25$, and the confidence threshold is set to $0.05$. In the segmentation part, the results from the two segmentation heads are merged, and the output is upsampled from $384 \times 640$ to $1080 \times 1920$.

\section{Experimental Results}

\subsection{Environment Setup}
We conducted our experiments using 8 Nvidia V100 GPUs for training. PyTorch 1.10~\cite{pytorch} and TensorFlow 2.8.0~\cite{tensorflow} were used to implement our models and training pipeline, while OpenCV 4.6.0~\cite{opencv_library} was used for image pre-processing.

Our model architecture was based on the publicly available PyTorch implementations of YOLOP~\cite{wu2022yolop} and YOLOv7~\cite{yolov7}. To migrate the model from PyTorch to TensorFlow, we first translated the PyTorch model into ONNX\footnote{https://onnx.ai/} format, and then used the onnx2tflite\footnote{https://github.com/MPolaris/onnx2tflite} toolkit to convert ONNX into TensorFlow (.h5) and TFLite model (.tflite).

\subsection{Main Results} 
We present the performance of our model on the final testing dataset provided by the contest organizer at different training stages. Initially, we trained the model only on the BDD100K~\cite{Yu2018BDD100KAD} dataset. However, due to the variation in the data distribution between BDD100K and the target task, the model may not be able to generalize well on the target task.

To address this issue, we added the iVS dataset to the training process and performed mix data finetuning (i.e. the third stage). This approach enabled the model to adapt itself to better fit the target task, as the iVS dataset provided additional data with a similar data distribution to the target task. By training on this diverse dataset, the model was able to learn more effectively from the data and improve its performance on the target task.

The performance of our proposed model is evaluated through various training stages. In the pretraining without mosaic stage, as depicted in  Table~\ref{table:val}, the model is trained on BDD100K dataset, which effectively boosts the performance of all.

Based on YOLOv4~\cite{Bochkovskiy2020YOLOv4OS}, we integrate mosaic technology in our model training. However, in the pretraining stage with mosaic shown in Table~\ref{table:val}, we notice a decrease in performance across all tasks. The implementation of the mosaic technique does not yield improved performance, which could potentially be attributed to its training exclusively on the BDD100K dataset. As a result, the model may be more suited to the BDD100K dataset, leading to a slight decline in performance when applied to the iVS dataset. Nevertheless, further finetuning on the iVS dataset enables the model to achieve enhanced performance.

In the third stage, the model is finetuned using a mix of the BDD100K and iVS datasets with mosaic augmentation, which resulted in a significant improvement in object detection and lane line segmentation performance. Additionally, in the last 10 epochs, the mosaic augmentation was turned off to allow the model to recover its adaptability to normal images.

\begin{table}[ht!]
\centering
\caption{The test performance on the iVS dataset provided by the contest organizer.}
\fontsize{8.5pt}{10pt}\selectfont
\begin{tabular}{c|ccc}
Model                   & \begin{tabular}{@{}c@{}}Object \\ Detection \\ (mAP@0.5) \end{tabular} &  \begin{tabular}{@{}c@{}}Drivable Area \\ Segmentation \\ (mIoU) \end{tabular}  &  \begin{tabular}{@{}c@{}}lane Line \\ Segmentation \\ (mIoU) \end{tabular}   \\ 
\hline
Pretraining w/o mosaic  & 0.445    &  0.837  &  0.433  \\
Pretraining w/ mosaic   & 0.417    &  0.852  &  0.379  \\
Finetuning              & 0.531    &  0.841  &  0.435  \\
\end{tabular}
\label{table:val}
\end{table}

\subsection{Testing Results in the Competition}

Table~\ref{table:final_res_public} shows the testing results of public dataset in the competition provided by the contest organizer. Our approach is effective for both object detection and segmentation tasks, achieving 0.495 mAP and 0.401 mIoU on pretraining with mosaic stage. Finetuning the model on the mix dataset improved the performance to 0.540 mAP and 0.615 mIoU, demonstrating the importance of the mix dataset in overcoming domain shift. Applying QAT to the finetuned model not only maintained the model's performance but also improved the detection task, which achieved 0.622 mAP and 0.612 mIoU.

\begin{table}[ht!]
\centering
\caption{The test performance on the final public testing dataset provided by the contest organizer.}
\fontsize{9.5pt}{12pt}\selectfont
\begin{tabular}{c|cc}
Model    & \begin{tabular}{@{}c@{}}Object Detection \\ (mAP@0.5) \end{tabular}  & \begin{tabular}{@{}c@{}}Segmentation \\ (mIoU) \end{tabular} \\ 
\hline
Pretraining w/ mosaic   & 0.495            & 0.401         \\
Finetuning              & 0.540            & 0.615         \\
QAT                     & 0.622            & 0.612
\end{tabular}
\label{table:final_res_public}
\end{table}

The testing results of private dataset in the competition provided by the contest organizer is shown in Table~\ref{table:final_res_private}. Our approach achieves state-of-the-art performance in both object detection and segmentation tasks, with 0.421 mAP and 0.612 mIoU.

\vspace{-0.15cm}
\begin{table}[ht!]
\centering
\caption{The test performance on the final private testing dataset provided by the contest organizer.}
\fontsize{9.5pt}{12pt}\selectfont
\begin{tabular}{c|cc}
\begin{tabular}{@{}c@{}}Object Detection \\ (mAP@0.5) \end{tabular}  & \begin{tabular}{@{}c@{}}Segmentation \\ (mIoU) \end{tabular} \\ 
\hline
0.421            & 0.612         \\
\end{tabular}
\label{table:final_res_private}
\end{table}

Moreover, Table~\ref{table:final_eff} shows that our quantization strategy effectively reduced the model size by 4 times and improved inference speed by 3 times. These results demonstrate the effectiveness of our quantization strategy not only in improving model performance but also in reducing computational cost and memory footprint, which is important for real-world deployment of deep learning models.

\begin{table}[ht!]
\centering
\caption{The comparison of 8-bits integer (INT8) weights and 32-bits floating (FP32) point weights. The model efficiency is conducted on MediaTek Dimensity 9200 Series Platform.}
\fontsize{10pt}{12pt}\selectfont
\begin{tabular}{c|ccc}
Model & Mode Size (M) & Power (mW) & Speed (us) \\
\hline
FP32  & 130.55        & 2279.0     & 87644.3    \\
INT8  & 31.79         & 2041.0     & 27086.1
\end{tabular}
\label{table:final_eff}
\end{table}

\vspace{-0.15cm}
\subsection{Quantization Strategy}

The performance of the quantized network using different quantization paradigms is presented in Table \ref{table:quant}. We first observe that Post-Training Quantization led to a significant performance drop in the segmentation tasks, with only 0.285 and 0.248 mIoU achieved for drivable area and lane line segmentation, respectively.

However, this performance drop can be mitigated by adopting a Quantization-Aware Training (QAT) strategy. Our experimental results demonstrate the effectiveness of QAT in mitigating the performance drop caused by quantization. Specifically, the quantized network achieved an 0.569 mAP for object detection and 0.852 mIoU for drivable area segmentation and 0.402 mIoU for lane line segmentation.

These findings demonstrate the effectiveness of the QAT strategy in boosting the performance of quantized network, as compared to the Post-Training Quantization strategy.

\vspace{-0.2cm}
\begin{table}[ht!]
\centering
\caption{The test performance of model after three-stage training with different quantization paradigms on the iVS dataset provided by the contest organizer.}
\fontsize{9.5pt}{12pt}\selectfont
\begin{tabular}{c|ccc}
Model                   & \begin{tabular}{@{}c@{}}Object \\ Detection \\ (mAP@0.5) \end{tabular} &  \begin{tabular}{@{}c@{}}Drivable Area \\ Segmentation \\ (mIoU) \end{tabular}  &  \begin{tabular}{@{}c@{}}lane Line \\ Segmentation \\ (mIoU) \end{tabular}   \\ 
\hline
original (fp32)       & 0.582    &   0.842   &   0.397  \\
PTQ (int8)            & 0.557    &   0.285   &   0.248  \\
QAT (int8)            & 0.569    &  0.852    &  0.402
\end{tabular}
\label{table:quant}
\end{table}

\section{Conclusion}
In this work, we have successfully implemented a light-weighted object detection and segmentation model. To improve its efficiency, we explored the effectiveness of two techniques: quantization-aware training and mix data finetuning (i.e. the third stage). Through extensive experimentation, we have demonstrated the effectiveness of these techniques in improving the accuracy and efficiency of our model. Our final model has achieved competitive results on the target dataset, demonstrating its potential for real-world applications.

\bibliographystyle{IEEEbib}
\bibliography{icme2023template}

\end{document}